\documentclass[letterpaper,10pt,conference]{ieeeconf}
\IEEEoverridecommandlockouts       
\usepackage{xcolor}

\usepackage{graphicx}
\usepackage{float}
\usepackage{booktabs}
\usepackage{textcomp}
\usepackage{multirow}
\usepackage[table]{xcolor}

\usepackage{amsmath,amssymb,amsfonts}

\usepackage[caption=false,font=footnotesize]{subfig}

\usepackage{pgfplots}
\usepackage{pgfplotstable}
\usepackage{tikz}
\usetikzlibrary{calc}
\usepgfplotslibrary{groupplots}
\pgfplotsset{compat=1.18}

\usepackage{comment}
\usepackage{titlesec}
\titlespacing*{\section}{0pt}{6pt}{3pt}
\titlespacing*{\subsection}{0pt}{5pt}{2pt}
\titlespacing*{\subsubsection}{0pt}{4pt}{2pt}

\newcommand{\gy}[1]{{\color{blue}{#1}}}

\usepackage[
    pdfa,
    colorlinks,
    bookmarksopen,
    bookmarksnumbered,
    allcolors=blue
]{hyperref}

\title{\LARGE \bf SAFER-Nav: Enhancing Safety for Visual Robot Navigation via Segmentation-Aware
Fine-Tuning}

\definecolor{ricemediumblue}{HTML}{4D9AD4}

\author{
Geonyeong Ko$^{1,\dagger}$,
Giung Lee$^{2,\dagger}$,
and Changjoo Nam$^{1,3}$%
\thanks{$^{1}$Geonyeong Ko and Changjoo Nam are with the Department of
Electronic Engineering, Sogang University, Seoul, South Korea.
{\tt\footnotesize \{geonyeong, cjnam\}@sogang.ac.kr}}%
\thanks{$^{2}$Giung Lee is with the Department of Computer Science,
Rice University, Houston, TX, USA.
{\tt\footnotesize gl34@rice.edu}}%
\thanks{$^{3}$Changjoo Nam is also affiliated with Vertical Labs,
Co., Ltd., Seoul, South Korea.}%
\thanks{$^{\dagger}$Geonyeong Ko and Giung Lee contributed equally
to this work.}%
}

\begin{document}
\raggedbottom
  \maketitle
  \vspace{-12pt}
  \begin{abstract}
   Vision-based navigation models, particularly foundation models, generate viable trajectories from RGB observations alone. However, even state-of-the-art transformer- and diffusion-based policies struggle to generalize in unfamiliar deployment environments containing unseen obstacles or shifted conditions. The resulting trajectories often remain goal-directed but unsafe. Existing efforts improve safety through external trajectory correction or internal geometric priors, yet the resulting policies are not trained to explicitly represent obstacle boundaries or traversable free-space structure. To address this, we propose a navigation model that incorporates these structures directly into the policy via fine-tuning and is designed for transformer-based RGB navigation policies. Across three robot platforms, two indoor environments, and static and dynamic obstacle scenarios, our method reduces collisions per run from 1.76 to 0.20 and raises the goal arrival rate from 42\% to 93\% relative to ViNT, with consistent gains over NoMaD and their CARE-augmented variants. Project page: \nolinkurl{https://paper-demo.github.io/SAFER\_Nav/}



\end{abstract}


  \section{Introduction}

Recent progress in visual representation learning and large language models (LLMs) has accelerated
the development of vision-language and vision-based navigation systems with contextual 
reasoning and generalization abilities~\cite{song2023llm,zhou2024navgpt}. Goal-conditioned navigation models such as GNM~\cite{shah2023gnm},
ViNT~\cite{shah2023vint}, and NoMaD~\cite{sridhar2024nomad} enable zero-shot navigation across embodiments. However, reaching the goal does not
imply navigating safely. Recent studies show that even modern transformer- and diffusion-based
policies still suffer from frequent collisions in cluttered and unfamiliar environments
despite high navigation success rates~\cite{guerrier2026can, kim2025care}.

Recent safety-oriented methods either correct the predicted trajectory post-hoc with an external depth-based module~\cite{kim2025care} or add geometric priors to the policy~\cite{gode2025flownav}. In neither case is the policy itself trained to represent obstacle boundaries or traversable free space, leaving the safety signal external or purely geometric. We instead inject this structure into the policy, so that the policy becomes collision-aware through a semantic signal. To this end, we propose SAFER-Nav (\underline{S}egmentation-\underline{A}ware
\underline{F}ine-tuning for \underline{E}nhancing Safety of \underline{R}obot \underline{Nav}igation),
a segmentation-aware navigation model that injects segmentation-derived representations of obstacle regions and traversable free space directly into a transformer-based RGB navigation policy through fine-tuning.

Our contributions include: (i) A segmentation-aware design that makes safety a representation-level
property of an RGB navigation policy, injecting obstacle and free-space
structure into a frozen pretrained backbone through a small set of trainable
modules. (ii) (ii) A dual-pathway formulation in which an attention bridge refines the action
representation while an auxiliary branch, supervised by a safe target derived
from free-space structure, supplies the avoidance direction that ground-truth
trajectories do not contain. (iii) Real-world validation across multiple robot platforms with differing
camera characteristics, two indoor environments differing in illumination and
visual appearance, and static and dynamic obstacle scenarios. A controlled study
over all combinations of the injection points isolates which pathways produce
the gain, showing that internalizing safety as a representation-level feature
improves obstacle-aware navigation under unfamiliar deployment conditions.

  \section{Related Work}

\subsection{Visual and Vision-Language Navigation Models}

Goal-conditioned navigation foundation models such as GNM~\cite{shah2023gnm}, ViNT~\cite{shah2023vint}, and NoMaD~\cite{sridhar2024nomad}, trained on large heterogeneous datasets, generalize across environments and robot embodiments and provide a practical paradigm for RGB-based goal-conditioned navigation.

Language-conditioned and vision-language-action (VLA) navigation includes LeLaN~\cite{hirose2025lelan}, NaVILA~\cite{cheng2025navila}, OmniVLA~\cite{hirose2025omnivla}, and VAMOS~\cite{guamancastro2025vamos}. These works extend navigation beyond image-goal conditioning to support language instructions, multiple goal modalities, and embodiment-aware reasoning. We target the complementary problem of local physical safety within transformer-based RGB goal-conditioned policies.


\subsection{Semantic Segmentation and Traversability for Navigation}

Prior work estimates traversability from RGB observations using learned binary prediction, self-supervised visual features, and semantic representations~\cite{hirose2018gonet,frey23fast,aegidius2025watch}. Others incorporate semantic and traversability information into navigation through terrain segmentation and semantic local planning~\cite{guan2022ganav,roth2024viplanner}. Green for Go, Red for No~\cite{szvoren2026green} uses segmentation-based visual grounding for VLA navigation, EmbodiedDiffusion~\cite{zhura2026embodieddiffusion} couples traversability estimation with learned trajectory generation, and CrossTracer~\cite{wang2026crosstracer} refines VLA-generated navigation traces using traversability cues.


\subsection{Safety-Aware Visual Navigation}

CARE~\cite{kim2025care} estimates monocular depth~\cite{piccinelli2025unidepthv2}, builds a local obstacle map, and applies APF-based repulsion~\cite{agirrebeitia2005new} to the predicted trajectory. This post-hoc correction keeps safety external to the policy and relies on depth estimation. FlowNav~\cite{gode2025flownav} integrates monocular depth priors and conditional flow matching into the policy, but its safety signal is primarily geometric. Neither approach explicitly trains the policy to distinguish between traversable and non-traversable regions using segmentation supervision.

SAFER-Nav incorporates this distinction through segmentation-aware representations and collision-aware waypoint prediction.

  \section{Problem Definition}

We consider \emph{goal-conditioned mobile robot navigation} in zero-shot indoor environments. At each step, the policy receives a short history of RGB observations $o_{t-P:t}$, a goal image $o_s$, and an aligned history of binary segmentation maps $m_{t-P:t}$ that distinguish \emph{traversable} from \emph{non-traversable} regions. The policy is
\begin{equation}
    \pi_\theta : \left(o_{t-P:t},\, o_s,\, m_{t-P:t}\right) \mapsto \left(\hat{d}_t,\, \hat{a}_t\right),
\end{equation}
where $\hat{d}_t$ is the predicted temporal distance to the goal and $\hat{a}_t$ is a short-horizon waypoint sequence for local motion planning, parameterized as
\begin{equation}
    \small
    \hat{a}_t = \{\hat{w}_{t}^{(h)}\}_{h=1}^{H}, \;
    \hat{w}_{t}^{(h)} = (\Delta x^{(h)}, \Delta y^{(h)}, \sin\theta^{(h)}, \cos\theta^{(h)}),
\end{equation}
where $H$ is the waypoint horizon, and each waypoint encodes local position and heading in the local frame of the robot. A trajectory is \emph{safe} if its executed path remains within traversable regions.

\section{Proposed Method: SAFER-Nav}

\begin{figure*}[t]
    \centering
    \setlength{\abovecaptionskip}{-4pt}
    \includegraphics[width=0.90\linewidth]{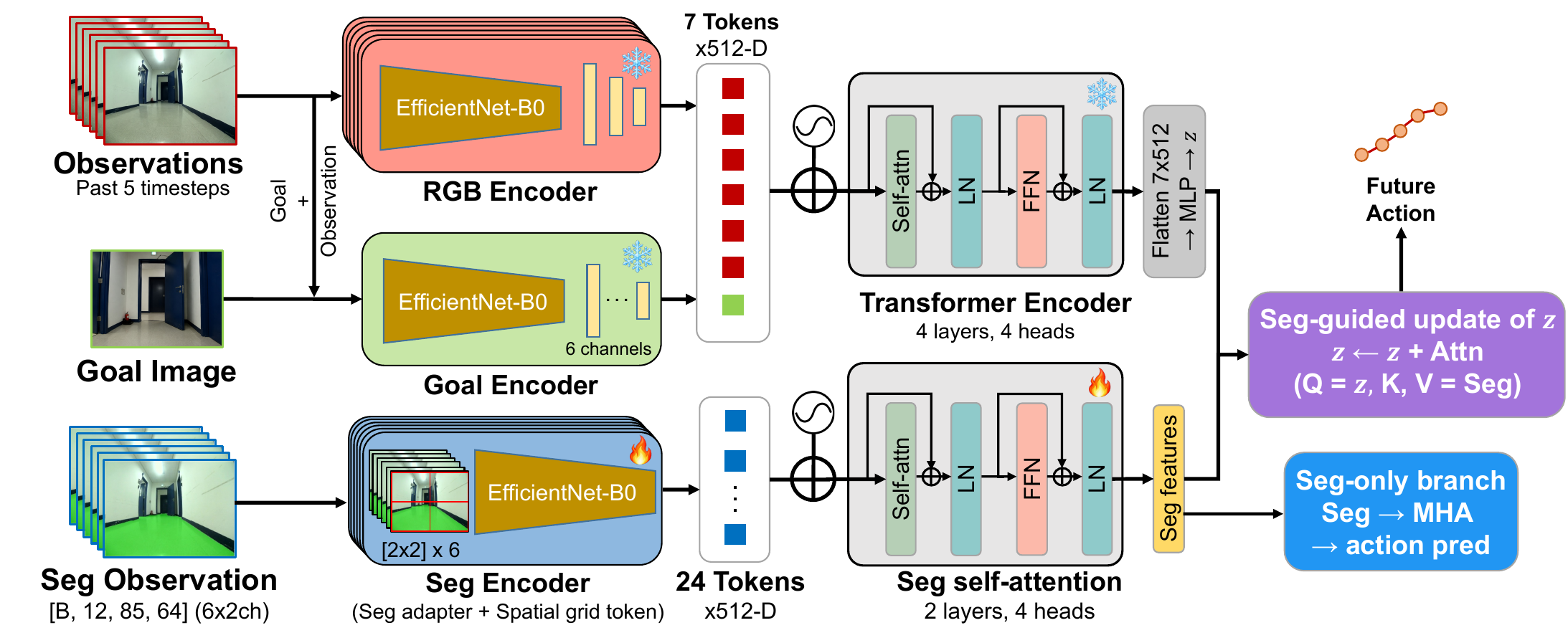}
    \caption{Overview of SAFER-Nav. A trainable segmentation branch augments a pretrained RGB-goal backbone to refine the latent representation and support safety-aware waypoint prediction. The snowflake icon indicates parameters that remain frozen during fine-tuning.}
    \label{fig1:SAFER-Nav_overview}\vspace{-15pt}
\end{figure*}

SAFER-Nav incorporates segmentation-aware safety information into a pretrained RGB-based navigation policy. As shown in Fig.~\ref{fig1:SAFER-Nav_overview}, it comprises an RGB-goal encoder, a segmentation encoder, a representation-level fusion module, and dual action prediction pathways. The RGB-goal branch follows a pretrained navigation backbone, producing a latent representation from the observation history and goal image. In parallel, a trainable segmentation branch converts aligned binary traversability masks into spatiotemporal tokens, refines them with a lightweight transformer, and injects them into the latent representation through attention-based fusion. The refined representation drives the main waypoint and distance heads, while an auxiliary segmentation-only branch predicts a safety-oriented waypoint sequence. At inference, the two predictions are dynamically blended.

\subsection{Input Representation}
The model takes three inputs: (i) a sequence of 6 RGB observations $o_{t-P:t}$ with $P=5$, each $85 \times 64 \times 3$; (ii) a goal image $o_s$ jointly encoded with the current observation following the early-fusion strategy of ViNT; and (iii) aligned binary segmentation masks $m_{t-P:t}$ identifying traversable floor regions. Each mask is a two-channel one-hot tensor, resized via nearest-neighbor interpolation to preserve sharp boundaries, giving a segmentation input of shape $[B, 12, 85, 64]$.

\paragraph{Segmentation labels:}
Training masks are generated offline from RGB frames using OneFormer~\cite{jain2023oneformer}, a panoptic segmentation model pretrained on ADE20K~\cite{zhou2017scene}. Floor-related classes (ADE20K indices 3, 13, and 29) are merged into a single binary mask of traversable regions; the complement forms the non-traversable region. The same model produces masks online at deployment, so no additional human annotation or fine-tuning is required.

\subsection{Network Architecture}

\paragraph{RGB-goal branch}
The RGB-goal branch follows the ViNT architecture. An EfficientNet-B0~\cite{tan2019efficientnet} observation encoder $\psi$ tokenizes each of the 6 observation frames into a 512-dimensional embedding $\psi(o_i) \in \mathbb{R}^{512}$. A separate EfficientNet-B0 goal-fusion encoder $\phi$ jointly encodes the current observation and goal image via channel-wise concatenation, producing $\phi(o_t, o_s) \in \mathbb{R}^{512}$. Both encoders are initialized from the pretrained ViNT checkpoint and frozen during training.

\paragraph{Segmentation branch}
The segmentation branch consists of an adapter, an encoder, a spatial token grid, and a self-attention block. A $1 \times 1$ convolution ($2 \rightarrow 3$ channels) maps the two-channel mask to three channels for EfficientNet-B0 input. A separate EfficientNet-B0, initialized from the observation encoder's weights, then encodes the adapted mask. Unlike the RGB branch, the segmentation encoder retains the $2 \times 2$ spatial structure of the final feature map, yielding 4 tokens per frame, which are projected from 1280 to 512 dimensions for $6 \times 4 = 24$ total tokens. Each token receives two additive positional encodings: a sinusoidal temporal encoding shared across the 4 spatial positions within a frame, and a learnable spatial embedding shared across all 6 frames. The 24 tokens are then refined by a 2-layer pre-norm transformer encoder with GELU activations and a final layer normalization.

\paragraph{Decoder and prediction heads}
The 6 observation tokens and the goal token pass through a transformer encoder $f$ with 4 layers and 4 attention heads, compared with ViNT's 2 heads. We flatten the output and pass it through a ReLU MLP ($3584 \rightarrow 512 \rightarrow 256 \rightarrow 128 \rightarrow 64 \rightarrow 32$) to obtain $\mathbf{z} \in \mathbb{R}^{32}$. After the segmentation-based refinement described below, a linear head ($32 \rightarrow 1$) predicts temporal distance $d$, while a 3-layer MLP ($32 \rightarrow 64 \rightarrow 128 \rightarrow 20$) with layer normalization, ReLU, and dropout ($p=0.1$) predicts $H=5$ waypoints $\hat{a} \in \mathbb{R}^{5 \times 4}$. Each waypoint contains robot-frame $(\Delta x, \Delta y, \sin\theta, \cos\theta)$. We cumulatively sum the predicted position increments and $\ell_2$-normalize the heading vectors.

\subsection{Segmentation-to-Action Mechanisms}
\label{sec:mechanisms}
\begin{figure}[tbp]
    \centering
    \setlength{\abovecaptionskip}{-2pt}
    \setlength{\belowcaptionskip}{-6pt}
    \includegraphics[width=1.0\linewidth]{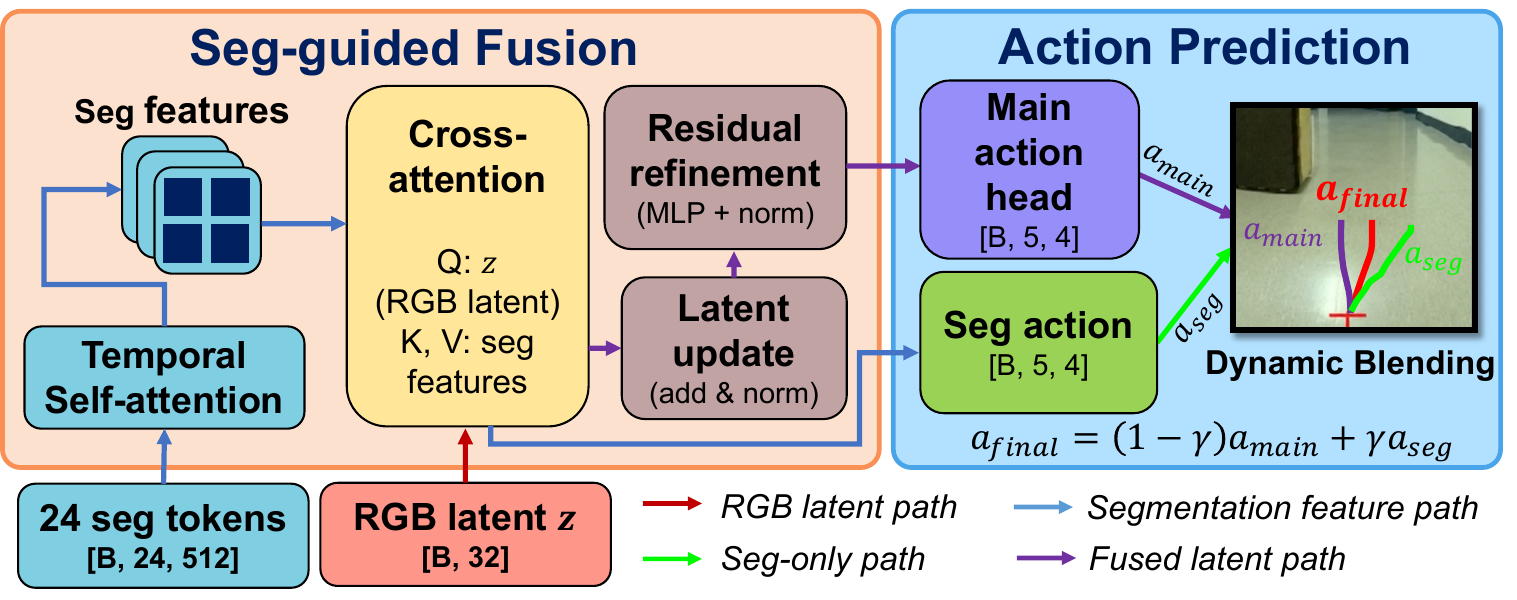}
    \caption{Seg-guided fusion detail. Segmentation features refine the RGB latent representation and support an auxiliary segmentation-only action branch for safety-aware waypoint prediction.}
    \label{fig2:RGB_Seg_path_details}\vspace{-15pt}
\end{figure}

As illustrated in Fig.~\ref{fig2:RGB_Seg_path_details}, segmentation information influences the policy through two complementary pathways at different stages of the pipeline. Segmentation can in principle be injected at three distinct points: \emph{before} the RGB transformer, by fusing segmentation tokens into the observation tokens; \emph{after} the transformer, by refining the decoded representation $\mathbf{z}$; and \emph{outside} the main pathway, through an auxiliary branch that predicts its own action. We implemented all three and selected the final design empirically; the combination study in Sec.~\ref{sec:combination} reports the full comparison. The two pathways retained in SAFER-Nav are described below, while the rejected pre-transformer variant is described in Sec.~\ref{sec:combination}.

\paragraph{Direct path (after the encoder)}
After the encoder produces $\mathbf{z} \in \mathbb{R}^{32}$, a residual attention bridge refines it over the segmentation tokens. The representation $\mathbf{z}$ is projected to 512 dimensions and used as a single query attending over the 24 segmentation tokens; the attended vector is projected back to 32 dimensions and added to $\mathbf{z}$ through a residual connection with layer normalization. This update is applied before the main prediction head, enabling late-stage refinement of the action representation given the current obstacle layout.

\paragraph{Segmentation-only action branch (independent)}
A learnable temporal query $\mathbf{q}_{\mathrm{seg}} \in \mathbb{R}^{512}$ attends over the 24 segmentation tokens to produce a pooled representation. A separate MLP ($512 \rightarrow 256 \rightarrow 128 \rightarrow 20$), with layer normalization, ReLU, and dropout ($p=0.1$), decodes this into an auxiliary waypoint sequence $\hat{a}_{\mathrm{seg}} \in \mathbb{R}^{5 \times 4}$. This branch is supervised by a \emph{safe target} (defined below) rather than the ground-truth trajectory.

\subsection{Training Objective}
Our training objective follows the standard supervision used in visual navigation, combining a distance loss and an action loss, augmented with a safety-oriented loss for the segmentation-only branch:
\begin{equation}
\mathcal{L}_{\mathrm{total}} = \lambda_d \mathcal{L}_{\mathrm{dist}} + \lambda_a \mathcal{L}_{\mathrm{action}} + \lambda_s \mathcal{L}_{\mathrm{safe}}.
\end{equation}

\paragraph{Distance loss $\mathcal{L}_{\mathrm{dist}}$}
Mean squared error between the predicted and ground-truth temporal distance to the goal:
\begin{equation}
\mathcal{L}_{\mathrm{dist}} = \frac{1}{B}\sum_{i=1}^{B}\left(\hat{d}_i - d_i\right)^2,
\end{equation}
where $B$ is the batch size.

\paragraph{Perspective-weighted action loss $\mathcal{L}_{\mathrm{action}}$}
MSE between the main action prediction and the ground truth, scaled per sample by the obstacle density with perspective weighting that emphasizes near-field rows. The weight follows $w(y) = \exp(3.0 \cdot y / R)$, where $y$ is the row index and $R$ is the number of image rows, so bottom rows contribute roughly $20\times$ more than top rows when computing the perspective-weighted obstacle ratio $\rho \in [0, 1]$. The per-sample loss weight is $1 + 2\rho \in [1.0, 3.0]$.

\paragraph{Safe target loss $\mathcal{L}_{\mathrm{safe}}$}
The segmentation-only branch is supervised by a \emph{safe target} computed online, biasing the trajectory toward free space when obstacles are present. For each sample, the perspective-weighted free-space ratio is estimated on each lateral side of the full image width. Obstacles are detected separately by a multi-band analysis over 4 horizontal bands with per-band thresholds, restricted to the central 60\% so that off-path structure such as side walls is excluded. Far-field bands require obstacle persistence across at least 2 of the 6 frames. Let $\Delta_{\mathrm{free}}$ denote the left--right free-space difference. A dead zone applies when $|\Delta_{\mathrm{free}}| < 0.05$; otherwise, a lateral avoidance offset is
\begin{equation}
\Delta y_{\mathrm{avoid}} = -\,\operatorname{sign}(\Delta_{\mathrm{free}})\cdot \beta \cdot s_{\mathrm{speed}} \cdot c_{\mathrm{free}},
\end{equation}
where $\operatorname{sign}(\Delta_{\mathrm{free}})$ indicates the side with more free space, $\beta = 0.5$, $s_{\mathrm{speed}} = \operatorname{clamp}(v_{\mathrm{GT}} / 0.5,\, 0.5,\, 1.5)$ scales with the ground-truth forward speed, and $c_{\mathrm{free}} = \operatorname{clamp}(|\Delta_{\mathrm{free}}| / 0.3,\, 0.3,\, 1.0)$ reflects the free-space asymmetry. When an obstacle is detected within the central 30\% of the image, a separate confidence term overrides the lateral estimate. The safe target is $\hat{a}^{(h)}_{\mathrm{safe},\,y} = \hat{a}^{(h)}_{\mathrm{GT},\,y} + \Delta y_{\mathrm{avoid}}$, clamped to $[-0.8, 0.8]$, with the heading recomputed.

The safe target loss combines MSE against this target with a hinge-form directional margin,
\begin{equation}
\mathcal{L}_{\mathrm{dir}} = \frac{1}{H}\sum_{h=1}^{H}\max\!\big(0,\; \mu - \operatorname{sign}(\Delta y_{\mathrm{avoid}})\cdot \hat{a}^{(h)}_{\mathrm{seg},\,y}\big),
\end{equation}
weighted by $0.5$, where $\mu = 0.05$ is the margin. The term penalizes predictions whose lateral direction opposes the safe avoidance direction, and is active only when avoidance is required.

\subsection{Training Details}
Training is conducted on the HuRoN dataset~\cite{hirose2023sacson}, which contains 576 indoor trajectories from a fisheye camera and yields approximately 145k training samples after a 6-frame sliding window. Each trajectory is paired with binary traversability masks generated by OneFormer~\cite{jain2023oneformer}. The network is initialized from the pretrained ViNT checkpoint. The RGB observation encoder, the goal-fusion encoder, and the RGB transformer remain frozen. The segmentation encoder is initialized from the observation encoder's weights, the segmentation adapter uses Kaiming-uniform initialization, and the remaining new modules are initialized from scratch. All new modules are trained at a learning rate of $1 \times 10^{-4}$ for 30 epochs with a batch size of 64, using AdamW with cosine decay and a 3-epoch linear warmup. The loss weights are $\lambda_d = 0.005$, $\lambda_a = 0.5$, $\lambda_s = 0.3$. Training runs on a single NVIDIA RTX PRO 6000 GPU and takes about 3 hours. 

\subsection{Inference: Dynamic Action Blending}
At deployment, the main action $\hat{a}$ and the segmentation-only avoidance action $\hat{a}_{\mathrm{seg}}$ are predicted at each step and blended as $\hat{a}_{\mathrm{final}} = (1 - \gamma)\,\hat{a} + \gamma\,\hat{a}_{\mathrm{seg}}$, where the weight $\gamma$ depends on the current obstacle layout. The instantaneous weight is
\begin{equation}
\gamma_{\mathrm{now}} =
\begin{cases}
\gamma_{\max} & \text{if multi-band detected,}\\
\operatorname{clamp}(\eta\rho,0,\gamma_{\max}) & \text{otherwise}
\end{cases}
\end{equation}
with $\eta = 2.0$, $\gamma_{\max} = 0.3$, and $\rho$ the perspective-weighted obstacle ratio. Temporal hysteresis $\gamma = \max(\gamma_{\mathrm{now}},\, \delta \gamma_{\mathrm{prev}})$ with $\delta = 0.7$ prevents oscillation when obstacles transiently leave the field of view. When no obstacles are detected, $\gamma = 0$ and the output reduces to the main action.

The perspective exponent follows from camera geometry, as the low forward-facing mounts map image rows monotonically to distance.


\subsection{Deployment}
At each timestep, the live RGB frame is processed in parallel by OneFormer, producing a binary traversability mask, and by a context buffer maintaining the 6-frame observation history. The two streams are passed to SAFER-Nav, and the resulting main and auxiliary predictions are blended as described above. The selected waypoint is published to a PD controller that generates linear and angular velocity commands, following ViNT's deployment protocol.

Segmentation and policy inference run as separate processes communicating through non-blocking queues, so the two stages are pipelined rather than serialized. On the laptop GPU, segmentation takes approximately 100--150\,ms per frame and policy inference approximately 67\,ms, while the waypoint loop runs at the 4\,Hz control rate of ViNT's deployment protocol. Both stages therefore complete within the 250\,ms control period. The decoupled design also allows the segmentation backbone to be exchanged for a lighter one without altering the policy, as the backbone-swap experiment in Sec.~\ref{sec:seg_source} indicates.

  \section{Experiment}
\subsection{Experimental Setup}


We deploy SAFER-Nav on three mobile robot configurations with different visual characteristics: a DJI RoboMaster S1 with a 120$^\circ$ field-of-view (FOV) camera, a TurtleBot4 with an Intel RealSense D435 RGB camera (69.4$^\circ$ FOV), and a LoCoBot with a 170$^\circ$ fisheye camera. Our LoCoBot configuration uses a TurtleBot4 mobile base and a custom 3D-printed mount that positions the fisheye camera to match the intended LoCoBot setup. The TurtleBot4 and LoCoBot configurations therefore share a mobile base but differ in camera placement and imaging characteristics.

For RoboMaster S1, TurtleBot4, and LoCoBot, respectively, the RGB streams published over ROS~2 have resolutions of $640 \times 360$, $320 \times 200$, and $320 \times 240$ pixels. Their camera optical centers are 240, 245, and 340\,mm above the ground, with longitudinal offsets of $+70$, $-60$, and $+10$\,mm from the robot geometric center; positive offsets indicate placement ahead of the center. SAFER-Nav inputs are resized to $85 \times 64$ pixels.

All robots operate under the same motion limits: 0.2\,m/s linear and 0.8\,rad/s angular velocity. The models run on a Lenovo ThinkPad P16 Gen 2 (Intel i7-14700HX, NVIDIA RTX 4090 Laptop, 16\,GB VRAM). SAFER-Nav evaluations use online OneFormer predictions with the same segmentation model and class mapping across platforms, without manually corrected masks.

\begin{figure}[h]
    \centering
    \setlength{\abovecaptionskip}{-5pt}
    \includegraphics[width=1.0\linewidth]{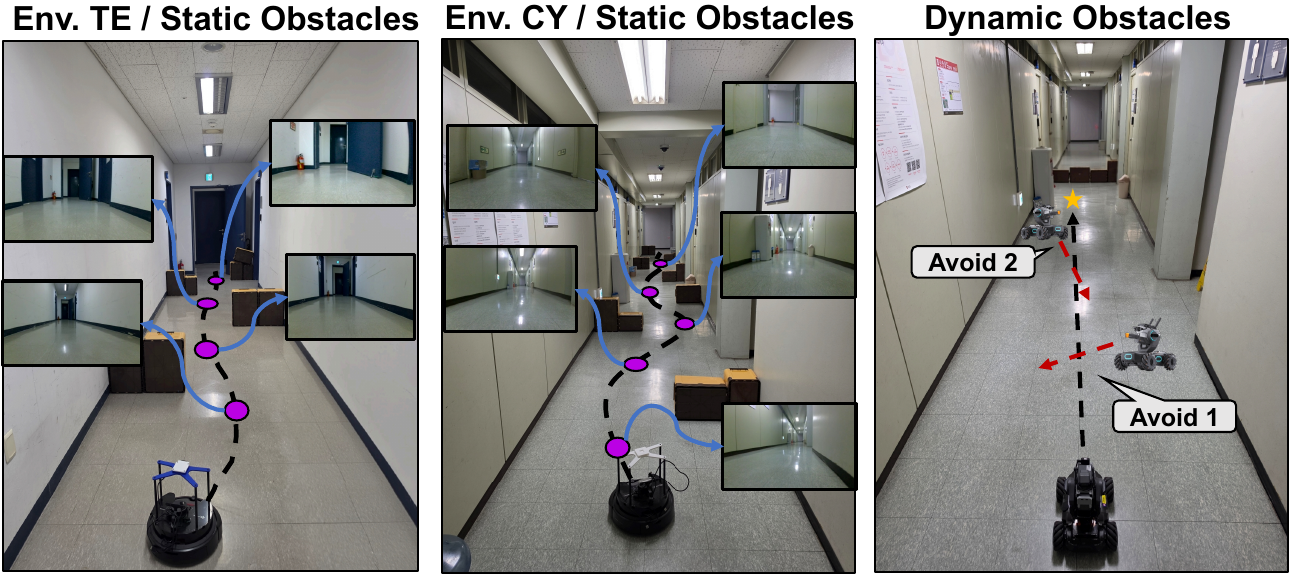}
    \caption{Overview of environments. Left and middle: static-obstacle setups in TE and CY, with maps and representative visual observations. Right: dynamic-obstacle setting, where the robot must continuously avoid the moving robot.}
    \label{fig3:Exp_Env_Figure}\vspace{-8pt}
\end{figure}

We compare SAFER-Nav against four baselines (ViNT, NoMaD, ViNT + CARE, NoMaD + CARE) in environments with static and dynamic obstacles (Fig.~\ref{fig3:Exp_Env_Figure}). ViNT and NoMaD serve as pretrained RGB goal-conditioned baselines, while their CARE variants allow comparison with post-hoc safety correction under the same sensing assumptions. Before introducing out-of-distribution (OOD) obstacles, we verified that all baselines could complete the routes using the same topological maps. SAFER-Nav was not tuned for individual robots or environments: its safe-target and blending constants were fixed once during development and held identical across all three cameras, both environments, and every trial. For static obstacles, we report path length and completion time (lower is more efficient), collision count per run (lower is safer), and arrival rate (higher is better). For dynamic obstacles, we focus on collision rate, as the question is whether the policy maintains safe local behavior under moving interference.



\paragraph{Static environment} We test all methods in two unseen indoor environments, denoted TE and CY, in different buildings and differing in lighting, floor and wall appearance, and geometry. TE is a 15.3 m corridor, 1.78 m wide, with two 90° right turns separated by a straight segment holding eight OOD plastic-box obstacles, ending in a narrow doorway. CY is 24.5 m, 1.7 m wide, with a 90° right turn, a straight segment holding 13 OOD obstacles, and a 90° left turn. Our method is built by fine-tuning ViNT, whose standard deployment constructs a topological map over the corridor. Following this setup, we place OOD obstacles at varied, challenging locations along each route and keep their positions fixed across methods and repeated trials for a controlled comparison. The robot must reach the goal by following the graph while avoiding the inserted obstacles.

\paragraph{Dynamic environment} In CY, we test whether the learned safety behavior extends beyond static clutter by replacing the plastic boxes with a moving mobile robot. Because conventional navigation datasets contain no robot-to-robot encounters, this setup evaluates dynamic obstacle avoidance and OOD generalization simultaneously. The mobile robot continuously crosses the path of the tested robot from arbitrary directions to cover diverse interaction patterns.

In both settings, we count a collision as any physical contact between the robot and an obstacle, including side-swipes rather than only head-on impacts. TurtleBot4 and LoCoBot share a bumper-equipped base that halts and reverses automatically on contact, so each triggered bumper event is counted as one collision. The RoboMaster S1 has no bumper, so an experimenter positioned in the environment counts contacts visually, using displacement of the plastic boxes as a corroborating cue. Repeated contacts during a single continuous interaction with the same obstacle are counted once. The same procedure is applied to every method evaluated on a given platform.

\subsection{Navigation with Static Obstacles}

Table~\ref{tab:static_envAB} summarizes static-obstacle results over 10 trials per method, platform, and environment. Across the successful trials, SAFER-Nav achieves the best safety–performance tradeoff on all three platforms, with the highest arrival rate and lowest collision frequency. SAFER-Nav reaches the goal at a higher rate than every baseline in both TE and CY on every platform. The baselines, by contrast, shift by up to 40 percentage points between environments. 
In TE, SAFER-Nav reaches the goal in 90\% of trials with zero collisions on RoboMaster, and maintains similarly strong performance (arrival rate / avg. collision) on TurtleBot4 (90\% / 0.24) and LoCoBot (100\% / 0.4). The baselines either collide more often or fail to reach the goal reliably. Notably, both ViNT+CARE and NoMaD+CARE underperform their base models on LoCoBot, suggesting that CARE's external depth-based correction is less well matched to its wider fisheye camera. In CY, SAFER-Nav achieves 90\% / 0.22 on RoboMaster, 90\% / 0.22 on TurtleBot4, and 100\% / 0.1 on LoCoBot, while baselines either reach the goal less reliably or make more collisions. Fig.~\ref{fig:static_goal_collision_scatter} plots goal arrival rate against collision count for all methods, platforms, and environments.

The gains in safety do not come at the cost of efficiency. In TE, the travel distance and completion time of SAFER-Nav stay comparable to ViNT across all platforms, indicating that the reduction in collisions is not achieved through conservative stopping or excessive detours. By contrast, NoMaD inflates both metrics. SAFER-Nav reduces travel distance by up to 39\% and completion time by up to 27\% relative to NoMaD, because when confronted with unseen obstacles NoMaD repeatedly rotates to search for a feasible subgoal direction and may continue making side-contact collisions before recovering. This is especially costly here, where the robot must avoid multiple OOD obstacles in sequence rather than perform a single correction.
\begin{figure}[t]
\centering
\footnotesize
\setlength{\abovecaptionskip}{-3pt}
\scalebox{0.93}{
\begin{tikzpicture}[trim left=0pt]

\begin{scope}[xshift=0.5cm]
\begin{groupplot}[
    group style={
        group size=3 by 2,
        horizontal sep=0.6cm,
        vertical sep=0.7cm
    },
    width=0.22\textwidth,
    height=0.16\textheight,
    xmin=-0.1, xmax=7.2,
    ymin=0, ymax=105,
    grid=both,
    grid style={dotted},
    tick label style={font=\scriptsize},
    label style={font=\scriptsize},
    title style={font=\scriptsize, yshift=-6pt},
]

\nextgroupplot[title={RoboMaster (TE)}]
\addplot[ only marks, mark=star, mark size=2.5pt, blue, line width=1.5pt] coordinates {(0,90)};
\addplot[only marks, mark=square*, red] coordinates {(3.17,60)};
\addplot[only marks, mark=triangle*, orange!90!black] coordinates {(3.67,30)};
\addplot[only marks, mark=diamond*, teal!70!black] coordinates {(1.4,50)};
\addplot[only marks, mark=*, purple] coordinates {(4.0,20)};

\nextgroupplot[title={TurtleBot4 (TE)}]
\addplot[ only marks, mark=star, mark size=2.5pt, blue, line width=1.5pt] coordinates {(0.24,90)};
\addplot[only marks, mark=square*, red] coordinates {(3.0,30)};
\addplot[only marks, mark=triangle*, orange!90!black] coordinates {(7.0,10)};
\addplot[only marks, mark=diamond*, teal!70!black] coordinates {(3.25,40)};
\addplot[only marks, mark=*, purple] coordinates {(4.5,20)};

\nextgroupplot[title={LoCoBot (TE)}]
\addplot[ only marks, mark=star, mark size=2.5pt, blue, line width=1.5pt] coordinates {(0.4,100)};
\addplot[only marks, mark=square, red, mark size=2.5pt] coordinates {(2.5,20)};
\addplot[only marks, mark=triangle*, orange!90!black, mark size=1.8pt] coordinates {(2.5,20)};

\nextgroupplot[title={RoboMaster (CY)}]
\addplot[ only marks, mark=star, mark size=2.5pt, blue, line width=1.5pt] coordinates {(0.22,90)};
\addplot[only marks, mark=square*, red] coordinates {(0.75,40)};
\addplot[only marks, mark=triangle*, orange!90!black] coordinates {(3.4,50)};
\addplot[only marks, mark=diamond*, teal!70!black] coordinates {(1.0,10)};
\addplot[only marks, mark=*, purple] coordinates {(2.0,10)};

\nextgroupplot[title={TurtleBot4 (CY)}]
\addplot[ only marks, mark=star, mark size=2.5pt, blue, line width=1.5pt] coordinates {(0.22,90)};
\addplot[only marks, mark=square*, red] coordinates {(1.0,40)};
\addplot[only marks, mark=triangle*, orange!90!black] coordinates {(4.5,20)};
\addplot[only marks, mark=diamond*, teal!70!black] coordinates {(3.33,30)};
\addplot[only marks, mark=*, purple] coordinates {(1.5,20)};

\nextgroupplot[title={LoCoBot (CY)}]
\addplot[ only marks, mark=star, mark size=2.5pt, blue, line width=1.5pt] coordinates {(0.1,100)};
\addplot[only marks, mark=square*, red] coordinates {(0.17,60)};
\addplot[only marks, mark=triangle*, orange!90!black] coordinates {(4.75,40)};
\addplot[only marks, mark=diamond*, teal!70!black] coordinates {(1.25,40)};
\addplot[only marks, mark=*, purple] coordinates {(3.0,30)};

\end{groupplot}

\end{scope}

\node[anchor=north, font=\scriptsize] at (group c2r2.south) [yshift=-0.28cm] {
    Average collision counts per run
};

\node[anchor=south, font=\scriptsize, rotate=90]
     at ($(group c1r1.west)!0.5!(group c1r2.west)+(-0.4cm,0)$)
     {Goal arrival rate (\%)};

\node[anchor=south, font=\scriptsize, draw=black!55, line width=0.4pt,
      inner sep=3pt, rounded corners=1pt]
     at ($(group c2r1.north)+(0,0.45cm)$) {%
    \tikz[baseline=-0.5ex]{
        \draw[blue, line width=1.5pt, line cap=round]
        plot[mark=star, mark size=2pt] coordinates {(0,0)};
    }~SAFER-Nav (Ours)\quad
    {\color{red}$\blacksquare$}~ViNT\quad
    {\color{orange!90!black}$\blacktriangle$}~NoMaD\quad
    {\color{teal!70!black}$\blacklozenge$}~ViNT+CARE\quad
    {\color{purple}\large$\bullet$}~NoMaD+CARE
};
\end{tikzpicture}
}
\caption{Scatter plot of static-obstacle results: $x$: average collisions per run; $y$: goal arrival rate (\%). Methods closer to the top-left corner perform better overall, where SAFER-Nav dominates the top-left region across all platforms and environments.
}
\label{fig:static_goal_collision_scatter}\vspace{-18pt}
\end{figure}

\begin{table*}[t]
\small
\centering
\setlength{\abovecaptionskip}{0pt}
\caption{Comparison of navigation performance under static obstacle environments.
Metrics: goal arrival rate (\%), collisions per run, distance (m), and completion time (s). N/A: no successful trial.}
\label{tab:static_envAB}

\setlength{\tabcolsep}{1.2pt}
\renewcommand{\arraystretch}{0.95}

\scalebox{0.83}{
\begin{tabular}{llcccccccc}
\toprule
&&
\multicolumn{4}{c}{\textbf{Environment TE}} &
\multicolumn{4}{c}{\textbf{Environment CY}} \\
\cmidrule(lr){3-6}
\cmidrule(lr){7-10}

\textbf{Robot} &
\textbf{Model} &
\textbf{Goal\%} &
\textbf{\#Coll.} &
\textbf{Dist. (m)} &
\textbf{Time (s)} &
\textbf{Goal\%} &
\textbf{\#Coll.} &
\textbf{Dist. (m)} &
\textbf{Time (s)} \\
\midrule

\multirow{5}{*}{\textbf{RoboMaster}}
& \cellcolor{green!15}SAFER-Nav (Ours)
& \cellcolor{green!15}\textbf{0.9}
& \cellcolor{green!15}\textbf{0}
& \cellcolor{green!15}14.87 $\pm$ 0.49
& \cellcolor{green!15}77.91 $\pm$ 7.26
& \cellcolor{green!15}\textbf{0.9}
& \cellcolor{green!15}\textbf{0.22}
& \cellcolor{green!15}24.98 $\pm$ 0.44
& \cellcolor{green!15}130.84 $\pm$ 5.67 \\

& ViNT
& 0.6 & 3.17
& 14.59 $\pm$ 0.67
& 79.99 $\pm$ 6.95
& 0.4 & 0.75
& 26.20 $\pm$ 0.65
& 128.63 $\pm$ 3.12 \\

& NoMaD
& 0.3 & 3.67
& 16.38 $\pm$ 0.20
& 87.50 $\pm$ 5.72
& 0.5 & 3.4
& 25.48 $\pm$ 0.78
& 122.27 $\pm$ 3.70 \\

& ViNT + CARE
& 0.5 & 1.4
& 15.17 $\pm$ 0.62
& 82.98 $\pm$ 4.22
& 0.1 & 1
& 26.35
& 138.253 \\

& NoMaD + CARE
& 0.2 & 4
& 17.07 $\pm$ 1.09
& 91.75 $\pm$ 3.18
& 0.1 & 2
& 25.974
& 131.193 \\

\midrule

\multirow{5}{*}{\textbf{TurtleBot4}}
& \cellcolor{green!15}SAFER-Nav (Ours)
& \cellcolor{green!15}\textbf{0.9}
& \cellcolor{green!15}\textbf{0.24}
& \cellcolor{green!15}14.41 $\pm$ 1.09
& \cellcolor{green!15}87.65 $\pm$ 7.74
& \cellcolor{green!15}\textbf{0.9}
& \cellcolor{green!15}\textbf{0.22}
& \cellcolor{green!15}25.26 $\pm$ 0.74
& \cellcolor{green!15}137.51 $\pm$ 3.09 \\

& ViNT
& 0.3 & 3
& 15.21 $\pm$ 0.52
& 84.03 $\pm$ 2.28
& 0.4 & 1
& 26.12 $\pm$ 0.73
& 141.89 $\pm$ 6.90 \\

& NoMaD
& 0.1 & 7
& 23.421
& 120.503
& 0.2 & 4.5
& 26.33 $\pm$ 0.15
& 125.73 $\pm$ 5.49 \\

& ViNT + CARE
& 0.4 & 3.25
& 14.35 $\pm$ 0.28
& 85.75 $\pm$ 4.60
& 0.3 & 3.33
& 26.84 $\pm$ 0.28
& 134.37 $\pm$ 3.16 \\

& NoMaD + CARE
& 0.2 & 4.5
& 17.28 $\pm$ 2.31
& 107.75 $\pm$ 7.07
& 0.2 & 1.5
& 26.08 $\pm$ 0.17
& 147.95 $\pm$ 6.16 \\

\midrule

\multirow{5}{*}{\textbf{LoCoBot}}
& \cellcolor{green!15}SAFER-Nav (Ours)
& \cellcolor{green!15}\textbf{1.0}
& \cellcolor{green!15}\textbf{0.4}
& \cellcolor{green!15}14.97 $\pm$ 0.34
& \cellcolor{green!15}85.13 $\pm$ 4.33
& \cellcolor{green!15}\textbf{1.0}
& \cellcolor{green!15}\textbf{0.1}
& \cellcolor{green!15}25.75 $\pm$ 0.91
& \cellcolor{green!15}134.63 $\pm$ 2.42 \\

& ViNT
& 0.2 & 2.5
& 17.18 $\pm$ 0.21
& 82.19 $\pm$ 2.74
& 0.6 & 0.17
& 25.88 $\pm$ 0.82
& 126.31 $\pm$ 6.09 \\

& NoMaD
& 0.2 & 2.5
& 18.78 $\pm$ 0.22
& 102.34 $\pm$ 1.23
& 0.4 & 4.75
& 26.14 $\pm$ 0.27
& 129.94 $\pm$ 5.83 \\

& ViNT + CARE
& 0 & N/A
& N/A
& N/A
& 0.4 & 1.25
& 25.29 $\pm$ 0.14
& 142.72 $\pm$ 4.70 \\

& NoMaD + CARE
& 0 & N/A
& N/A
& N/A
& 0.3 & 3
& 25.92 $\pm$ 0.49
& 152.59 $\pm$ 2.18 \\

\bottomrule
\end{tabular}
}
\vspace{-15pt}
\end{table*}

\subsection{Navigation with Dynamic Obstacles}


We evaluate SAFER-Nav in CY with a dynamic obstacle (a teleoperated RoboMaster) to assess whether learned safety behavior transfers beyond static clutter. Unlike a walking human, whose large image footprint makes them easy to detect, the moving robot presents a smaller and OOD visual target relative to the training data, posing a more demanding test of obstacle awareness. We deploy SAFER-Nav on RoboMaster and evaluate two scenarios: (i) \textit{Corner-appear}, in which the TurtleBot4 enters the field of view from a side corridor, and (ii) \textit{Front-approach}, in which it approaches head-on and stops in its path.


We run 10 trials per method per scenario (Table~\ref{tab:dynamic}). ViNT and NoMaD collide in most trials, with near-complete failure in front-approach. CARE-augmented baselines reduce collisions substantially in front-approach, likely because depth-based repulsion detects the approaching robot once it occupies sufficient image area. They still collide in corner-appear, however, where the obstacle enters abruptly from the side and may not be captured early enough to trigger repulsion. SAFER-Nav avoids all collisions, indicating that segmentation-aware traversability representations provide reliable dynamic obstacle awareness even for small, OOD targets.
\vspace{-7pt}
\begin{table}[h]
\small
\centering
\setlength{\abovecaptionskip}{-2pt}
\caption{Number of trials with collisions (out of 10 trials) for each dynamic-obstacle scenario.}
\label{tab:dynamic}
\scalebox{0.9}{
\begin{tabular}{lcc}
\toprule
\textbf{Model} & \textbf{(i) Corner-appear} & \textbf{(ii) Front-approach} \\
\midrule
\rowcolor{green!15} SAFER-Nav (Ours) & \textbf{0/10} & \textbf{0/10} \\
ViNT         & 5/10 & 9/10 \\
NoMaD        & 6/10 & 8/10 \\
ViNT + CARE  & 3/10 & 2/10 \\
NoMaD + CARE & 2/10 & 2/10 \\
\bottomrule
\end{tabular}
}\vspace{-15pt}
\end{table}

\subsection{Ablation Study}

\subsubsection{Combination Study and Architecture Selection}
\label{sec:combination}


We compare seven non-empty combinations of three independently switchable mechanisms: (a) \textbf{pre-transformer fusion}, which updates RGB observation tokens through cross-attention to segmentation tokens before the transformer; (b) the \textbf{post-transformer bridge}, which refines the decoded representation $\mathbf{z}$; and (c) the \textbf{auxiliary branch}, which predicts an action supervised by the safe target (Sec.~\ref{sec:mechanisms}). All configurations use the same HuRoN training split, frozen RGB backbone, optimizer, learning rate, batch size, and training schedule. Because every configuration receives identical task adaptation on identical data, differences among C1--C7 reflect the segmentation pathway rather than fine-tuning itself. The all-off configuration is the segmentation-free backbone, represented by the pretrained ViNT baseline in the real-robot experiments.


\begin{figure*}[t]
    \centering
    \setlength{\abovecaptionskip}{-5pt}
    \includegraphics[width=0.85\textwidth]{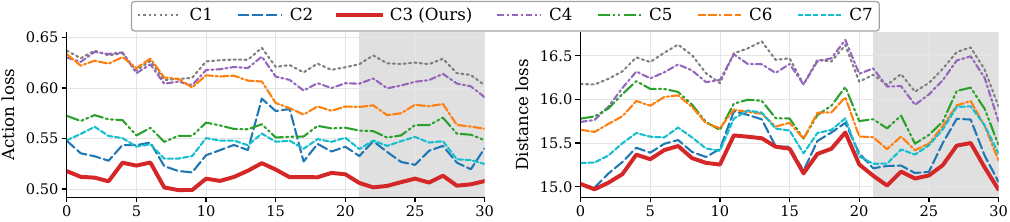}
    \caption{Learning curves over the 30 training epochs (horizontal axis) for the seven non-empty segmentation-to-action configurations C1--C7 of Table~\ref{tab:combination}. The shaded region marks epochs 21--30, the readout window used for the summary statistics in that table. Curves are drawn with a centered 3-epoch moving average, as per-epoch values are computed from a single batch; the reported statistics are computed from the unsmoothed values. Lower values are better for both the action and temporal-distance losses.}
    \label{fig:combination_curves}
    \vspace{-20pt}
\end{figure*}


Table~\ref{tab:combination} reports perspective-weighted action loss and temporal-distance loss on the held-out HuRoN validation split, averaged over epochs 21--30. We train each configuration once and evaluate one batch per epoch, so the standard deviations describe fluctuations across epochs within a run rather than variability across random seeds. Fig.~\ref{fig:combination_curves} shows the learning curves, while the table reports statistics from unsmoothed values. Distance loss rises as action loss decreases, so we use it only to compare configurations.

\begin{table}[t]
\small
\centering
\setlength{\abovecaptionskip}{0pt}
\caption{Combination study: (a) pre-transformer fusion, (b) post-transformer bridge, and (c) auxiliary branch. Held-out HuRoN losses are reported as mean $\pm$ std over epochs 21--30; lower is better.}
\label{tab:combination}
\setlength{\tabcolsep}{4pt}
\scalebox{0.9}{
\begin{tabular}{lccccc}
\toprule
& \textbf{(a)} & \textbf{(b)} & \textbf{(c)} & \textbf{Action loss} & \textbf{Dist. loss} \\
\midrule
C1 & --        & --        & \checkmark & 0.622 $\pm$ 0.016 & 16.28 $\pm$ 0.32 \\
C2 & --        & \checkmark & --        & 0.533 $\pm$ 0.024 & 15.36 $\pm$ 0.39 \\
\rowcolor{green!15} C3 (Ours) & -- & \checkmark & \checkmark & \textbf{0.505 $\pm$ 0.013} & \textbf{15.19 $\pm$ 0.31} \\
C4 & \checkmark & --        & --        & 0.605 $\pm$ 0.013 & 16.18 $\pm$ 0.34 \\
C5 & \checkmark & --        & \checkmark & 0.557 $\pm$ 0.015 & 15.77 $\pm$ 0.40 \\
C6 & \checkmark & \checkmark & --        & 0.575 $\pm$ 0.018 & 15.60 $\pm$ 0.34 \\
C7 & \checkmark & \checkmark & \checkmark & 0.541 $\pm$ 0.021 & 15.54 $\pm$ 0.33 \\
\bottomrule
\end{tabular}
}\vspace{-18pt}
\end{table}

The post-transformer bridge and auxiliary branch each reduce action loss in all three available matched comparisons. Adding the bridge lowers the loss from C1 to C3, C4 to C6, and C5 to C7; adding the auxiliary branch lowers it from C2 to C3, C4 to C5, and C6 to C7. C3 achieves an action loss of 0.505, compared with 0.622 for the auxiliary-only configuration C1 and 0.533 for the bridge-only configuration C2. These comparisons support retaining both pathways.

The effect of pre-transformer fusion depends on which other mechanisms are present. It reduces action loss when added to the auxiliary-only configuration (C1 to C5: 0.622 to 0.557), but increases it when the bridge is present (C2 to C6: 0.533 to 0.575; C3 to C7: 0.505 to 0.541). C3 also has a lower distance loss than C7, which uses all three mechanisms (15.19 versus 15.54). Early fusion therefore provides no action-loss benefit in the tested configurations containing the bridge. Since all seven configurations share the same training setup and differ only in how segmentation reaches the action, the spread between the best and worst configuration (0.505 versus 0.622, a 19\% gap) is produced by the segmentation pathway alone, with the contribution of fine-tuning held constant across every run. We select C3, which has the lowest values on both metrics, as SAFER-Nav for the main navigation experiments and evaluate its two pathways separately on hardware below.

\subsubsection{Segmentation-to-Action Pathway Ablation}
To relate these offline results to deployed behavior, we compare the selected C3 architecture (direct path + seg-only branch) against the two variants that retain only one mechanism, C1 and C2.

With only the seg-only branch, the model consistently fails to avoid obstacles and exhibits frequent head-on collisions. The segmentation tokens have no residual connection to $\mathbf{z}$, so the obstacle layout captured by the branch cannot refine the main waypoint prediction, and the auxiliary avoidance signal is too weak on its own to compensate through blending. In TE, this variant succeeds in only 1 of 10 trials.

With only the direct path, the robot begins to deviate from straight-line trajectories in the presence of obstacles, indicating that late-stage attention over the segmentation tokens introduces obstacle awareness into $\mathbf{z}$. However, the avoidance maneuvers remain inconsistent in magnitude and direction; this variant succeeds in 3 of 10 trials in TE, an improvement over the seg-only branch but still insufficient. Without the safe-target supervision from the seg-only branch, the direct path modulates the representation but has no training signal specifying \emph{where} to steer, producing hesitant or insufficient lateral corrections.

Reliable collision avoidance and consistent goal reaching emerge only when both mechanisms operate together. The direct path injects the current obstacle layout into $\mathbf{z}$ through a residual attention bridge, ensuring obstacle awareness in the main action representation. The seg-only branch, supervised by the safe-target loss, provides a directionally consistent avoidance signal of sufficient magnitude. The two pathways are complementary: the direct path determines \emph{that} avoidance is needed, while the seg-only branch determines \emph{how} to avoid.

\subsection{Robustness to the Segmentation Source}
\label{sec:seg_source}
\label{sec:seg_robustness}

\paragraph{Robustness to the segmentation backbone}
To assess dependence on the segmentation model, we replace OneFormer with SegFormer~\cite{xie2021segformer} and Mask2Former~\cite{cheng2022masked} at inference while keeping the trained SAFER-Nav (C3) weights fixed. All three models use ADE20K-pretrained checkpoints~\cite{zhou2017scene} without fine-tuning and share the same traversability class mapping. We run 10 trials per backbone on the LoCoBot static-obstacle route in TE. OneFormer and SegFormer each achieve 10/10 goal arrivals, while Mask2Former achieves 9/10 (Table~\ref{tab:seg_backbone}). The policy thus accommodates these backbones without retraining, with similar observed goal-reaching performance on this route.

\begin{table}[h]
\small
\centering
\setlength{\abovecaptionskip}{0pt}
\caption{Goal arrival on LoCoBot in TE with different segmentation backbones and fixed SAFER-Nav (C3) weights. Each backbone is evaluated over 10 trials.}
\label{tab:seg_backbone}
\setlength{\tabcolsep}{6pt}
\begin{tabular}{lc}
\toprule
\textbf{Segmentation backbone} & \textbf{Arrival rate} \\
\midrule
OneFormer~\cite{jain2023oneformer} (default) & 10/10 \\
Mask2Former~\cite{cheng2022masked} & 9/10 \\
SegFormer~\cite{xie2021segformer} & 10/10 \\
\bottomrule
\end{tabular}
\vspace{-15pt}
\end{table}

\begin{figure}[h]
    \centering
    \setlength{\abovecaptionskip}{-3pt}
    \includegraphics[width=\linewidth]{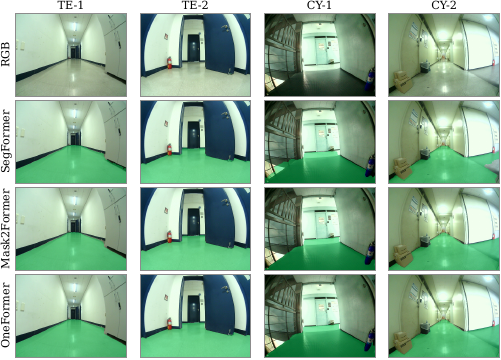}
    \caption{Online traversability predictions from three segmentation backbones on identical frames from TE and CY, showing differences at object boundaries and around floor reflections.}
    \label{fig:seg_backbone_qual}\vspace{-8pt}
\end{figure}

Fig.~\ref{fig:seg_backbone_qual} compares predictions on four frames from TE and CY. The models differ mainly at object boundaries and around specular reflections on polished floors, but identify broadly similar traversable regions. The corresponding traversable pixel fractions differ by less than 2 percentage points across backbones on each frame (Table~\ref{tab:seg_floor_ratio}). These fractions describe the overall extent of predicted free space, but do not establish agreement at individual obstacle boundaries.
\vspace{-13pt}
\begin{table}[h]
\small
\centering
\setlength{\abovecaptionskip}{0pt}
\caption{Traversable pixel fractions for the frames in Fig.~\ref{fig:seg_backbone_qual} Max.\ diff.\ is the range across backbones in percentage points.}
\label{tab:seg_floor_ratio}
\setlength{\tabcolsep}{4pt}
\begin{tabular}{lcccc}
\toprule
\textbf{Scene} & \textbf{SegFormer} & \textbf{Mask2Former} & \textbf{OneFormer} & \textbf{Max.\ diff.} \\
\midrule
TE-1 & 32.9\% & 32.1\% & 32.4\% & 0.8\%p \\
TE-2 & 32.7\% & 32.2\% & 32.1\% & 0.6\%p \\
CY-1 & 27.0\% & 25.6\% & 25.8\% & 1.4\%p \\
CY-2 & 27.6\% & 29.1\% & 29.4\% & 1.8\%p \\
\bottomrule
\end{tabular}
\vspace{-8pt}
\end{table}


\paragraph{Segmentation errors and their limited effect}
During deployment, we observe boundary flicker on glossy floors, missed thin or low-contrast obstacle edges, and occasional single-frame dropouts during rapid scene changes. We did not identify any of these errors as the cause of a collision in our trials. Three mechanisms bound their effect: the policy receives a six-frame mask history; the far-field bands used in safe-target computation require obstacle evidence in at least two frames; and temporal hysteresis, $\gamma = \max(\gamma_{\mathrm{now}}, \delta\gamma_{\mathrm{prev}})$ with $\delta = 0.7$, reduces the auxiliary contribution gradually when obstacle evidence disappears. Persistent errors can still affect both the main prediction through the bridge and the auxiliary action. In the limiting case where segmentation reports the scene as entirely traversable, $\rho$ falls to zero, driving $\gamma$ to zero, and the output reduces to the base policy prediction by construction.

  \section{Conclusion}

We presented SAFER-Nav, a segmentation-aware fine-tuning framework that improves goal-reaching and collision avoidance in real-world visual navigation through complementary representation and action pathways. Evaluation remains limited to a ViNT-based backbone. Although the same blending rule and parameters are used across all experiments, this heuristic may produce overly conservative or delayed avoidance. Future work will extend the framework to other navigation backbones and learn the intervention policy.





  {\footnotesize \bibliography{ref}}

\begin{thebibliography}{10}
\providecommand{\url}[1]{#1}
\csname url@samestyle\endcsname
\providecommand{\newblock}{\relax}
\providecommand{\bibinfo}[2]{#2}
\providecommand{\BIBentrySTDinterwordspacing}{\spaceskip=0pt\relax}
\providecommand{\BIBentryALTinterwordstretchfactor}{4}
\providecommand{\BIBentryALTinterwordspacing}{\spaceskip=\fontdimen2\font plus
\BIBentryALTinterwordstretchfactor\fontdimen3\font minus \fontdimen4\font\relax}
\providecommand{\BIBforeignlanguage}[2]{{%
\expandafter\ifx\csname l@#1\endcsname\relax
\typeout{** WARNING: IEEEtran.bst: No hyphenation pattern has been}%
\typeout{** loaded for the language `#1'. Using the pattern for}%
\typeout{** the default language instead.}%
\else
\language=\csname l@#1\endcsname
\fi
#2}}
\providecommand{\BIBdecl}{\relax}
\BIBdecl

\bibitem{song2023llm}
C.~H. Song, J.~Wu, C.~Washington, B.~M. Sadler, W.-L. Chao, and Y.~Su, ``{LLM-Planner: Few-shot grounded planning for embodied agents with large language models},'' in \emph{Proc. of the Int. Conf. Comp. Vis.}, 2023.

\bibitem{zhou2024navgpt}
G.~Zhou, Y.~Hong, and Q.~Wu, ``{NavGPT: Explicit reasoning in vision-and-language navigation with large language models},'' in \emph{Proc. of the AAAI Conf. on Art. Int.}, 2024.

\bibitem{shah2023gnm}
D.~Shah, A.~Sridhar, A.~Bhorkar, N.~Hirose, and S.~Levine, ``{GNM: A general navigation model to drive any robot},'' in \emph{Proc. of IEEE Int. Conf. on Robot. and Autom.}, 2023.

\bibitem{shah2023vint}
D.~Shah, A.~Sridhar, N.~Dashora, K.~Stachowicz, K.~Black, N.~Hirose, and S.~Levine, ``{ViNT: A Foundation Model for Visual Navigation},'' in \emph{Proc. of the Conf. on Robot Learning}, 2023.

\bibitem{sridhar2024nomad}
A.~Sridhar, D.~Shah, C.~Glossop, and S.~Levine, ``{NoMaD: Goal Masked Diffusion Policies for Navigation and Exploration},'' in \emph{Proc. of IEEE Int. Conf. on Robot. and Autom.}, 2024.

\bibitem{guerrier2026can}
M.~Guerrier, K.~Soma, J.~Pavlasek, and G.~Beltrame, ``{Can Vision Foundation Models Navigate? Zero-Shot Real-World Evaluation and Lessons Learned},'' \emph{arXiv:2603.25937}, 2026.

\bibitem{kim2025care}
J.~Kim, J.~Sim, W.~Kim, K.~Sycara, and C.~Nam, ``{CARE: Enhancing safety of visual navigation through collision avoidance via repulsive estimation},'' in \emph{Proc. of the Conf. on Robot Learning}, 2025.

\bibitem{gode2025flownav}
S.~Gode, A.~Nayak, D.~N. Oliveira, M.~Krawez, C.~Schmid, and W.~Burgard, ``{FlowNav: Combining Flow Matching and Depth Priors for Efficient Navigation},'' in \emph{Proc. of IEEE/RSJ Int. Conf. on Int. Robot. and Sys.}, 2025.

\bibitem{hirose2025lelan}
N.~Hirose, C.~Glossop, A.~Sridhar, O.~Mees, and S.~Levine, ``{LeLaN: Learning A Language-conditioned Navigation Policy from In-the-Wild Video},'' in \emph{Proc. of the Conf. on Robot Learning}, 2024.

\bibitem{cheng2025navila}
A.-C. Cheng, Y.~Ji, Z.~Yang, Z.~Gongye, X.~Zou, J.~Kautz, E.~Biyik, H.~Yin, S.~Liu, and X.~Wang, ``{NaVILA: Legged Robot Vision-Language-Action Model for Navigation},'' in \emph{Proc. of Robot.: Sci. and Sys.}, 2025.

\bibitem{hirose2025omnivla}
N.~Hirose, C.~Glossop, D.~Shah, and S.~Levine, ``{OmniVLA: An Omni-Modal Vision-Language-Action Model for Robot Navigation},'' arXiv:2509.19480, 2025.

\bibitem{guamancastro2025vamos}
M.~Guaman~Castro, S.~Rajagopal, D.~Gorbatov, M.~Schmittle, R.~Baijal, O.~Zhang, R.~Scalise, S.~Talia, E.~Romig, C.~de~Melo, B.~Boots, and A.~Gupta, ``{VAMOS: A Hierarchical Vision-Language-Action Model for Capability-Modulated and Steerable Navigation},'' 2025.

\bibitem{hirose2018gonet}
N.~Hirose, A.~Sadeghian, M.~V{\'a}zquez, P.~Goebel, and S.~Savarese, ``{GONet: A Semi-Supervised Deep Learning Approach For Traversability Estimation},'' in \emph{Proc. of IEEE/RSJ Int. Conf. on Int. Robot. and Sys.}, 2018.

\bibitem{frey23fast}
J.~Frey, M.~Mattamala, N.~Chebrolu, C.~Cadena, M.~Fallon, and M.~Hutter, ``{Fast Traversability Estimation for Wild Visual Navigation},'' in \emph{Proc. of Robot.: Sci. and Sys.}, 2023.

\bibitem{aegidius2025watch}
S.~{\AE}gidius, D.~Hadjivelichkov, J.~Jiao, J.~Embley-Riches, and D.~Kanoulas, ``{Watch Your STEPP: Semantic Traversability Estimation using Pose Projected Features},'' in \emph{Proc. of IEEE Int. Conf. on Robot. and Autom.}, 2025.

\bibitem{guan2022ganav}
T.~Guan, D.~Kothandaraman, R.~Chandra, A.~J. Sathyamoorthy, K.~Weerakoon, and D.~Manocha, ``{GA-Nav: Efficient terrain segmentation for robot navigation in unstructured outdoor environments},'' \emph{IEEE Robot. Autom. Lett.}, vol.~7, no.~3, 2022.

\bibitem{roth2024viplanner}
P.~Roth, J.~Nubert, F.~Yang, M.~Mittal, and M.~Hutter, ``{ViPlanner: Visual Semantic Imperative Learning for Local Navigation},'' in \emph{Proc. of IEEE Int. Conf. on Robot. and Autom.}, 2024.

\bibitem{szvoren2026green}
A.~Szvoren, D.~Kanoulas, and N.~Tuptuk, ``{Green for Go, Red for No: Visual Grounding via Semantic Segmentation for VLA Navigation Policies},'' \emph{arXiv:2607.05122}, 2026.

\bibitem{zhura2026embodieddiffusion}
I.~Zhura, S.~Karaf, F.~Batool, N.~D.~W. Mudalige, V.~Serpiva, A.~A. Abdulkarim, A.~Fedoseev, D.~Seyidov, H.~Amjad, and D.~Tsetserukou, ``{EmbodiedDiffusion: End-to-End Traversability-Guided Visual Diffusion for Heterogeneous Robot Navigation},'' \emph{IEEE Robot. Autom. Lett.}, 2026.

\bibitem{wang2026crosstracer}
Y.~Wang, S.~Wang, Z.~Sun, W.~Chi, L.~Lin, J.~Wang, and W.~Xu, ``{CrossTracer: Cross-Embodiment Navigation via VLA Model Reasoning and Trace Residuals Adapting},'' \emph{arXiv:2608.06688}, 2026.

\bibitem{piccinelli2025unidepthv2}
L.~Piccinelli, C.~Sakaridis, Y.-H. Yang, M.~Segu, S.~Li, W.~Abbeloos, and L.~Van~Gool, ``{UniDepthV2: Universal Monocular Metric Depth Estimation Made Simpler},'' \emph{IEEE Trans. Pattern Anal. Mach. Intell.}, 2025.

\bibitem{agirrebeitia2005new}
J.~Agirrebeitia, R.~Avil{\'e}s, I.~F. De~Bustos, and G.~Ajuria, ``{A new APF strategy for path planning in environments with obstacles},'' \emph{Mech. Mach. Theory}, vol.~40, no.~6, 2005.

\bibitem{jain2023oneformer}
J.~Jain, J.~Li, M.~T. Chiu, A.~Hassani, N.~Orlov, and H.~Shi, ``{OneFormer: One Transformer to Rule Universal Image Segmentation},'' in \emph{Proc. of the IEEE/CVF Conf. on Comp. Vis. and Patt. Recog.}, 2023.

\bibitem{zhou2017scene}
B.~Zhou, H.~Zhao, X.~Puig, S.~Fidler, A.~Barriuso, and A.~Torralba, ``{Scene Parsing through ADE20K Dataset},'' in \emph{Proc. of the IEEE/CVF Conf. on Comp. Vis. and Pattern Recog.}, 2017.

\bibitem{tan2019efficientnet}
M.~Tan and Q.~Le, ``{EfficientNet: Rethinking Model Scaling for Convolutional Neural Networks},'' in \emph{Proceedings of the 36th Int. Conf. on Machi. Learn.}, 2019.

\bibitem{hirose2023sacson}
N.~Hirose, D.~Shah, A.~Sridhar, and S.~Levine, ``{SACSoN: Scalable Autonomous Control for Social Navigation},'' \emph{IEEE Robot. Autom. Lett.}, vol.~9, no.~1, 2023.

\bibitem{xie2021segformer}
E.~Xie, W.~Wang, Z.~Yu, A.~Anandkumar, J.~M. Alvarez, and P.~Luo, ``{SegFormer: Simple and Efficient Design for Semantic Segmentation with Transformers},'' in \emph{Proc. of Adv. Neural Inf. Process. Syst.}, 2021.

\bibitem{cheng2022masked}
B.~Cheng, I.~Misra, A.~G. Schwing, A.~Kirillov, and R.~Girdhar, ``Masked-attention mask transformer for universal image segmentation,'' in \emph{Proc. of the IEEE/CVF Conf. on Comp. Vis. and Patt. Recog.}, 2022.

\end{thebibliography}

\end{document}